# NFDI4Health workflow and service for synthetic data generation, assessment and risk management

Sobhan MOAZEMI[a,1,2] Tim ADAMS[a,2] Hwei Geok NG[a] Lisa KÜHNEL[b,c] Julian SCHNEIDER[b] Anatol-Fiete NÄHER[d,e,f] Juliane FLUCK[b,g,h] and Holger FRÖHLICH[a,i,1]

[a] *Department of Bioinformatics, Fraunhofer Institute for Algorithms and Scientific Computing SCAI*
[b] *Knowledge Management, ZB MED – Information Centre for Life Sciences*
[c] *Graduate School DILS, Bielefeld Institute for Bioinformatics Infrastructure (BIBI), Faculty of Technology, Bielefeld University*
[d] *Digital Global Public Health, Hasso Plattner Institute for Digital Engineering, University of Potsdam*
[e] *Institute of Medical Informatics, Charité–Universitätsmedizin*
[f] *Method Development, Research Infrastructure, and Information Technology, Robert Koch Institute*
[g] *The Agricultural Faculty, University of Bonn*
[h] *Institute for Geodesy and Geoinformation, University of Bonn*
[i] *Bonn-Aachen International Center for Information Technology (B-it), University of Bonn*

**Abstract.** Individual health data is crucial for scientific advancements, particularly in developing Artificial Intelligence (AI); however, sharing real patient information is often restricted due to privacy concerns. A promising solution to this challenge is synthetic data generation. This technique creates entirely new datasets that mimic the statistical properties of real data, while preserving confidential patient information. In this paper, we present the workflow and different services developed in the context of Germany's National Data Infrastructure project NFDI4Health. First, two state-of-the-art AI tools (namely, VAMBN and MultiNODEs) for generating synthetic health data are outlined. Further, we introduce SYNDAT (a public web-based tool) which allows users to visualize and assess the quality and risk of synthetic data provided by desired generative models. Additionally, the utility of the proposed methods and the web-based tool is showcased using data from Alzheimer's Disease Neuroimaging Initiative (ADNI) and the Center for Cancer Registry Data of the Robert Koch Institute (RKI).

[1] Corresponding Author, Department of Bioinformatics, Fraunhofer Institute for Algorithms and Scientific Computing SCAI, Schloss Birlinghoven 1, 53757 Sankt Augustin, Germany; E-mail: sobhan.moazemi@scai.fraunhofer.de, holger.froehlich@scai.fraunhofer.de .
[2] equal contribution

1.  **Introduction**

Sharing real patient data from clinical studies is critical for scientific progress, but privacy regulations often restrict access. While federated machine learning is increasingly receiving attention, the implementation of according platforms poses non-trivial organizational, legal and technical challenges, including the existence of a common data model across different participating centers. One alternative solution is to generate synthetic datasets, which mimic the statistical properties of real data while limiting the release of confidential information.

In recent years, multiple approaches for generating synthetic data with the help of AI models have been published. For instance, generative models such as adversarial networks (GANs) [1] have demonstrated significant success in various (medical imaging) use-cases [2-4]. However, GANs frequently exhibit statistical mode collapse, raising concerns about the adequacy of synthetic data in capturing the true patient distributions. Furthermore, the generation of patient-level synthetic clinical study data poses unique challenges: a) Clinical studies often comprise a mixture of longitudinal and static data, and patient visits may have different time intervals. b) There are multiple data modalities comprising, e.g., continuous numerical, ordinal as well as categorical features. c) Missing values can occur at random as well as not at random, e.g. due to patient drop-out. These challenges require the development of specific methods that are tailored to address the aforementioned issues.

Our previously published Variational Autoencoder Bayesian Networks (VAMBN) [5], and Multimodal Neural Ordinary Differential Equations (MultiNODEs) [6] are aimed at providing realistic counterparts for clinical datasets while appropriately addressing the challenges mentioned above. To evaluate the quality of synthetic data, but also potential privacy risks we subsequently implemented the publicly available web-based tool, SYNDAT.

There have been efforts in related work to objectively assess quality of synthetic data provided by different generative AI tools and further to verify the utility of the generated data to solve a down-stream real-world problem. Arnold and Neunhoeffer [7] proposed a framework for the evaluation of the quality of differentially private synthesized data. Chen et al. [8] conducted a study to evaluate Synthea [9], a synthetic data generator leveraging a Generic Module Framework integrating a state transition mechanism, as applied for a colorectal cancer screening task. Although related works address some important aspects of synthetic data generation, to the best of our knowledge, there is no open-source tool available for researchers to objectively assess the quality and utility of synthetic data, particularly clinical patient data. To address this, SYNDAT publicly provides visualizations and evaluation tools for researchers for the objective assessment of the quality and utility of synthesized health data cohorts. Furthermore, with SYNDAT we aim at closing the gap of limited consensus when it comes to the systematic and objective comparison of generative models as identified by literature [7,8].

This paper provides a high-level overview about the NFDI4Health toolkit for synthetic health data generation and evaluation. The effectiveness of this approach is demonstrated using real-world datasets from the ADNI and the Cancer Registry Data of the RKI.

## 2. Methods

### 2.1. AI-Based Generation of Synthetic Clinical Study Data

To generate synthetic data in a way that addresses the specific challenges of multimodal longitudinal clinical study data, we have in the past developed two generative AI techniques: (1) VAMBN [5], and (2) MultiNODEs [6]. Briefly, VAMBN is a hybrid modeling approach, in which initially the potentially high dimensional feature space of a clinical study is grouped into modules. These modules could, e.g., represent demographics, volumes of defined brain regions, clinical assessments, biomarkers and others. Each of the modules is initially encoded in a lower dimensional space using a Heterogeneous Incomplete Variational Autoencoder [10], which accounts for missing values as well as different data modalities within each module. In a second step these modules are then connected via a Bayesian Network. At this point it is possible to model the sequence of different visits and to account for missingness not at random via specific auxiliary variables. An extension of VAMBN, named VAMBN Memorized Time Points (VAMBN-MT), has recently been introduced to improve the reconstruction of longitudinal dependencies via Long Short-Term Memory networks. More details are described in our previous papers [5,15].

MultiNODEs is an alternative method for handling mixed static and longitudinal data. The longitudinal part of the data is modeled on a continuous time scale using Neural Ordinary Differential Equations (NODE) [11]. More specifically, static as well as longitudinal data are initially encoded in a joint lower dimensional space, which represent the initial conditions of a NODE. Solving the NODE yields a latent disease trajectory, which can be decoded and mapped to the original data. Since MultiNODEs are trained via variational inference it is possible to generate highly realistic synthetic patient trajectories on continuous time scale, hence enabling interpolation between visits as well as extrapolation. A limitation of MultiNODEs compared to VAMBN is a significantly higher demand in terms of computational resources. Furthermore, VAMBN yields a "white box" model, which can be visualized as a graph. On the other hand, we found VAMBN to represent the correlation structure between variables less well than MultiNODEs [6]. All algorithms are available open source and can thus be employed within a data holding organization for data synthesization[2,3,4].

### 2.2. Evaluation of Synthetic Patient Data

Critical questions for any synthetic data generation approach are the realism of the generated data on the one hand and the potential risk to deduce characteristics of a real study participant from the synthetic data on the other hand. In order to address these points we have developed SYNDAT, a public web tool (https://syndat.scai.fraunhofer.de/) that helps researchers to evaluate synthetic data quality and privacy. Importantly, SYNDAT can also be downloaded and installed locally by a data holding organization to work with patient data internally. It is made available as an open source tool for common usage. To evaluate the quality of synthetic data, SYNDAT compares real and synthetic patient data by three quality metrics:

---

[2] https://github.com/nfdi4health/docker-vambn
[3] https://github.com/nfdi4health/vambn-extensions-evaluations
[4] https://github.com/SCAI-BIO/MultiNODEs

a) ability to discriminate between real and synthetic data by a machine learning (ML) classifier, trained based on a 5-fold cross validation data splits of the original and synthetic data. A low classifier performance is an indication of a high discrimination ability, b) similarity of marginal statistical distributions of each individual variable in real and synthetic datasets, measured via their Jensen-Shannon-Divergence (JSD), averaged over all available features and c) a score measuring the preservation of the correlation structure between variables in the synthetic data, which we compute based on the normalized quotient of the Frobenius norm of the pairwise feature correlation matrices of the real and synthetic data. We normalize each of the three metrics to a quality score in the range of 0-100 for a more intuitive understanding. Since a high score is in most disciplines associated with better performance, we inverse the JSD distance measure as well as the relative correlation difference such that a low value corresponds to a high score. Similarly, for the ML classifier, a bad performance (AUC close to 0.5) is converted to a high *Distinction Complexity* score.

SYNDAT also includes privacy risk scores based on the privacy framework of Giomi et al [5], which assess the risks of a) singling out a patient by rare or unique attributes (singling out risk), b) being able to link two sets of attributes of the same patient in external datasets A and B (linkability risk), and c) being able to deduce the value of a specific attribute in a given patient record (inference risk). In addition to viewing quality and security scores, SYNDAT allows users to visualize the data using t-SNE projections as well as plots of individual variables and correlation structures. In this context, SYNDAT also offers the possibility to automatically identify outliers in the synthetic data (see **Figure 1**). All of the above mentioned metrics and visualizations can be computed ad-hoc following a data upload in the associated dashboard and are available within minutes after triggering the processing procedure.

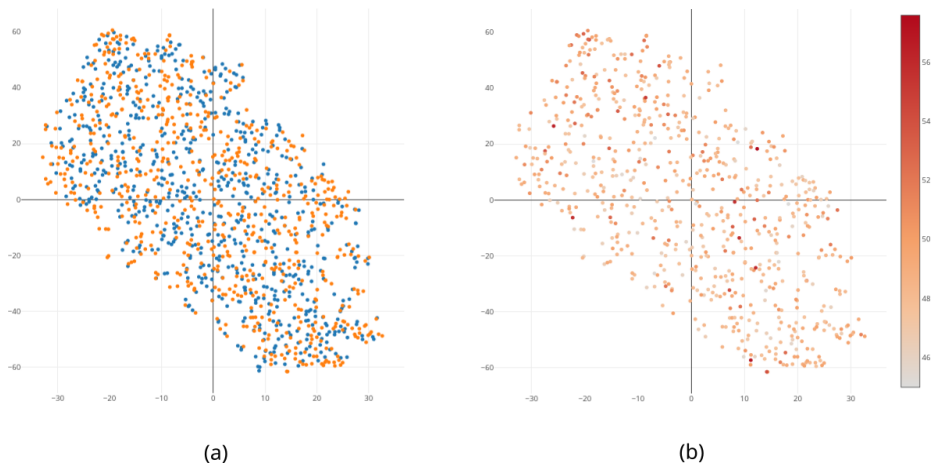

(a) (b)

**Figure 1.** Two-dimensional embedding of patient distribution of real and synthetic data (a) and computed outlier probabilities for synthetic data points (b) for the ADNI dataset.

*2.3.  Data*

We exemplify the use of VAMBN for synthetic data generation and SYNDAT for quality and risk assessment: In collaboration with the Center for German National Cancer Registry Data at the RKI [10], we created a synthetic data set using VAMBN based on oncology data of patients with glioblastoma multiforme [11]. The data is cross-sectional and comprises 9 variables with over 40,000 records.

The second dataset is from the Alzheimer's Disease Neuroimaging Initiative (ADNI[5]). It is a multimodal, longitudinal clinical study currently enrolling more than 2000 participants. To mitigate the high amount of missingness in consequent visits, we only evaluated baseline visits and subsequently removed rows with missing reports, resulting in a subset of 76 variables available for 665 patients.

## 3. Results

We found that synthetic data generated by VAMBN retains the statistical properties within variables as well as the pairwise correlation structures across variables for both tested datasets.. This even applies for clinical data that is inherently difficult to model due to difficulties such as missing and highly diverse data in terms of scales and properties. We highlight these capabilities based on the results we computed using SYNDAT. We achieved excellent results, especially for the RKI data set based on the metrics we introduced for SYNDAT. The synthetic data generated by VAMBN is nearly indistinguishable by a trained ML model. We computed this by the Discrimination Complexity score for SYNDAT in **Figure 2**. We found that the statistical distributions are kept almost identical to the real data with a Jensen-Shannon distance close to 0 corresponding to a Distribution Similarity score close to 100. We show exemplary distribution plots for two features in **Figure 3**. In addition to the distributions within each feature, the correlations across features were also preserved to a high degree as shown for two exemplary features in **Figure 4**. We achieved similar performance for the synthetic ADNI data for both the Distribution Similarity and feature Correlation Score, with a score of 87 for the JSD metric and a score of 80 for the pairwise correlation metric.

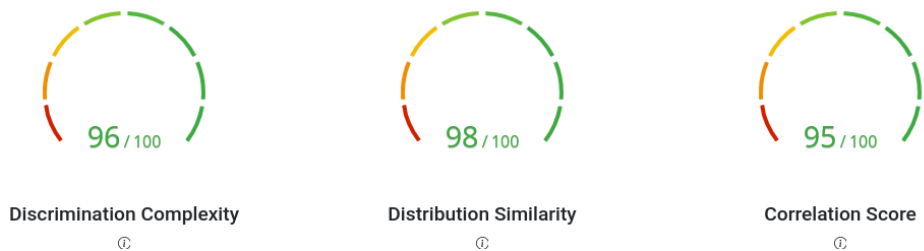

**Figure 2.** Exemplary VAMBN synthetic data quality results for the RKI data set as computed and displayed in the SYNDAT Dashboard.

---

[5] https://adni.loni.usc.edu

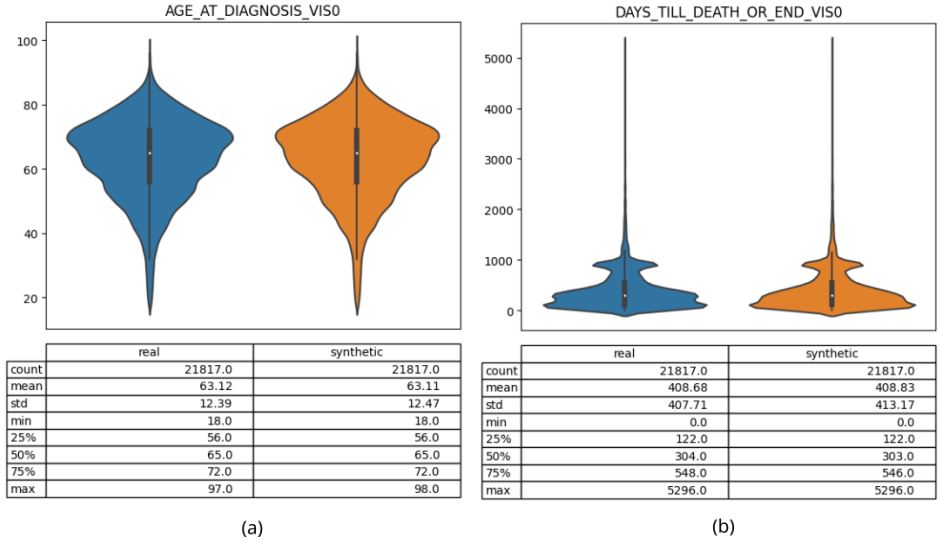

**Figure 3.** Visualization of two feature distributions for the RKI data for the age of diagnosis (a) and days until death or end of treatment (b).

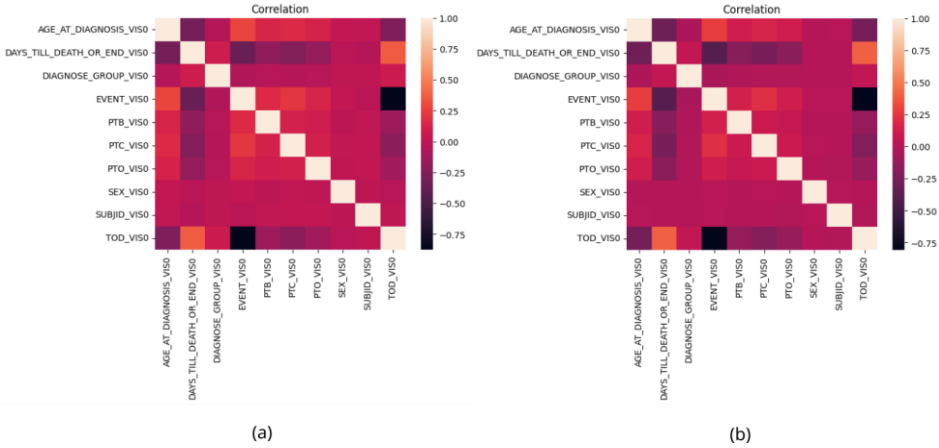

**Figure 4.** Pairwise feature correlations of the real (a) and synthetic (b) RKI data generated by VAMBN.

## 4. Discussion

Our contributions in the spectrum of NFDI4Health initiative address some of the crucial challenges in the field of federated machine learning in the extremely heterogeneous horizon of data-driven clinical patient studies. First, clinical patient cohorts often attribute inconsistent measurement frequencies with some patient-level static features alongside bed-side or lab values captured at often irregular time intervals. Further, patient data are multimodal in nature, as different information might be stored in different data types, e.g., ordinal, categorical, etc. Additionally, due to various reasons such as patient drop-out, missingness is an integral part of clinical data cohorts. Our previously developed generative models for synthetic health data, namely VAMBN and

MultiNODEs, have been successfully utilized to cope with these challenges in previous works [5,6,15].

Our findings suggest that using our proposed synthesization methods, clinical data can be modeled with high fidelity, preserving their original statistical properties as well as the correlations across features to a high degree.

A potential limitation of synthetic data as well as an ongoing topic of research is however the degree to which synthetic data can still be vulnerable to adversary attacks such as membership inference [16]. While adaptations to generative methods to implement concepts such as Differential Privacy (DP) aim to mitigate the risk of such attacks, it has been shown that data utility will generally decrease as privacy guarantees increase [17].

Another prominent approach of data sharing is the use of anonymization tools, such as ARX by Prasser et al. [18]. Similarly as for synthetic data, anonymized data will - with the degree depending on its implementation - always trade off data utility for privacy. It has also been shown that data that has been anonymized using paradigms
such as k-anonymity is still susceptible to privacy attacks such as Singling Out Attacks, where DP provides much better protection [19].

Given the importance of evaluating data utility and privacy in both synthetic and anonymized data, we aim to further improve on the proposed evaluation techniques to facilitate such assessments. Future research should also focus on the further development of implementations that provide privacy guarantees for generative methods while limiting potential loss of data utility.

As identified in the literature [7,8], another aspect which we aimed at addressing by introducing SYNDAT was the lack of consensus for the systematic and objective methods for comparing generative models, focusing primarily on synthetic health data cohorts. SYNDAT provides open-source tools for researchers and developers to objectively assess the quality and utility of synthetic data in terms of visualizations and evaluation metrics meticulously tailored for clinical patient data studies. Therefore, our approach keeps promise towards closing the gap of consensus by providing publicly available tools equipped with state-of-the-art objective measures.

## 5. Conclusion

NFDI4Health has the mission to facilitate the access to patient-level clinical study data in Germany. In this context synthetic data generation techniques play a relevant role, because these data can be shared with lower legal barriers than real patient data, hence fostering collaboration between organizations. Furthermore, data synthesization might be viewed as a potential alternative to more traditional anonymization techniques.

Synthesization of clinical study data comes along with unique challenges, which are addressed by the development of advanced and specifically tailored generative AI models (MultiNODEs, VAMBN, VAMBN-MT). Subsequently, SYNDAT allows quantifying the quality of synthetic data as well as potential privacy risks. Altogether, we are thus offering a solution, which supports the use of synthetic data generation approaches for clinical research and development.

Future developments could include the use of synthetic data to facilitate federated learning (FL), where data scientists need to develop algorithms without having direct access to data. Moreover, also learning of synthetic data over distributed datasets following the same common data model could be explored.

**Declarations**

*Study-Registry:* not applicable

*Conflict of Interest:* The authors declare no competing interests.

*Author contributions:* SM, TA, JF, HF: conceptualization of the study; SM, TA, HGN, LK, JS, and AFT: model development and data analysis; TA: SYNDAT development. All authors approved the manuscript in the submitted version and take responsibility for the scientific integrity of the work.

*Acknowledgement:* This work was done as part of the NFDI4Health Consortium (www.nfdi4health.de). We gratefully acknowledge the financial support of the Deutsche Forschungsgemeinschaft (DFG, German Research Foundation) - project number 442326535.

Data collection and sharing for this project was funded by the Alzheimer's Disease Neuroimaging Initiative (ADNI) (National Institutes of Health Grant U01 AG024904) and DOD ADNI (Department of Defense award number W81XWH-12-2-0012). ADNI is funded by the National Institute on Aging, the National Institute of Biomedical Imaging and Bioengineering, and through generous contributions from various organizations and companies. The Canadian Institutes of Health Research is providing funds to support ADNI clinical sites in Canada. Private sector contributions are facilitated by the Foundation for the National Institutes of Health (www.fnih.org). The grantee organization is the Northern California Institute for Research and Education, and the study is coordinated by the Alzheimer's Therapeutic Research Institute at the University of Southern California. ADNI data are disseminated by the Laboratory for Neuro Imaging at the University of Southern California. This research was also supported by NIH grants P30 AG010129 and K01 AG030514.

**References**

[1] Goodfellow I, Pouget-Abadie J, Mirza M, Xu B, Warde-Farley D, Ozair S, et al. Generative adversarial nets. In: Advances in neural information processing systems. 2014. p. 2672–80.
[2] Lei Y, Tian S, Zhu X, Han X, Wang B, He Y, et al. MRI-only based synthetic CT generation using dense cycle consistent generative adversarial networks. Med Phys. 2019;46:3565–81
[3] Yang G, Yu S, Dong H, Slabaugh G, Dragotti PL, Ye X, et al. DAGAN: Deep De-Aliasing Generative Adversarial Networks for fast compressed sensing MRI reconstruction. IEEE Trans Med Imaging. 2018;37:1310–21.
[4] Jordon J, Yoon J. PATE-GAN: generating synthetic data with differential privacy guarantees. In: International Conference on Learning Representations. 2019. p. 21.
[5] Gootjes-Dreesbach L, Sood M, Sahay A, Hofmann-Apitius M, Fröhlich H. (2020). Variational Autoencoder Modular Bayesian Networks for Simulation of Heterogeneous Clinical Study Data. Front Big Data. 2020 May 28;3:16. doi: 10.3389/fdata.2020.00016. PMID: 33693390; PMCID: PMC7931863.
[6] Wendland P, Birkenbihl C, Gomez-Freixa M, Sood M, Kschischo M, Fröhlich H. (2022). Generation of realistic synthetic data using Multimodal Neural Ordinary Differential Equations. Npj Digital Medicine 2022 5:1, 5(1), 1–10. https://doi.org/10.1038/s41746-022-00666-x
[7] Arnold C, Neunhoeffer M. Really Useful Synthetic Data -- A Framework to Evaluate the Quality of Differentially Private Synthetic Data. ArXiv, 2004.07740.
[8] Chen J, Chun D, Patel M, Chiang E, Tran D, Pratap A. The validity of synthetic clinical data: a validation study of a leading synthetic data generator (Synthea) using clinical quality measures. BMC Med Inform Decis Mak. 2019;19:44. Available from: https://doi.org/10.1186/s12911-019-0793-0.
[9] Walonoski J, Kramer M, Nichols J, Quina A, Moesel C, Hall D, Duffett C, Dube K, Gallagher T, McLachlan S. Synthea: An approach, method, and software mechanism for generating synthetic patients

and the synthetic electronic health care record. J Am Med Inform Assoc. 2018 Mar;25(3):230-238. doi: 10.1093/jamia/ocx079.
[10] Nazabal A, Olmos PM, Ghahramani Z, Valera I. Handling Incomplete Heterogeneous Data using VAEs [Internet]. arXiv; 2020 [cited 2024 Apr 22]. Available from: http://arxiv.org/abs/1807.03653
[11] Chen RTQ, Rubanova Y, Bettencourt J, Duvenaud D. Neural Ordinary Differential Equations [Internet]. arXiv; 2019 [cited 2024 Apr 22]. Available from: http://arxiv.org/abs/1806.07366
[12] Giomi M, Boenisch F, Wehmeyer C, Tasnád, B. (2023). A unified framework for quantifying privacy risk in synthetic data. PoPETs Proceedings, 2023(2), 312-328. https://doi.org/10.56553/popets-2023-0055
[13] Wolf U, Barnes B, Bertz J, Haberland J, Laudi A, Stöcker M, Schönfeld I, Kraywinkel K, Kurth BM. Das Zentrum für Krebsregisterdaten (ZfKD) im Robert Koch-Institut (RKI) in Berlin [The (German) Center for Cancer Registry Data (ZfKD) at the Robert Koch Institute (RKI) in Berlin]. Bundesgesundheitsblatt Gesundheitsforschung Gesundheitsschutz. 2011 Nov;54(11):1229-34. German. doi: 10.1007/s00103-011-1361-7. PMID: 22015795.
[14] Efremov L, Abera SF, Bedir A, Vordermark D, Medenwald D. Patterns of glioblastoma treatment and survival over a 16-years period: pooled data from the German Cancer Registries. J Cancer Res Clin Oncol. 2021 Nov;147(11):3381–90.
[15] Kühnel L, Schneider J, Perrar I, et al. Synthetic data generation for a longitudinal cohort study – evaluation, method extension and reproduction of published data analysis results. Sci Rep. 2024;14:14412. doi: 10.1038/s41598-024-62102-2.
[16] Zhang Z, Yan C, Malin BA. Membership inference attacks against synthetic health data. J Biomed Inform. 2022 Jan;125:103977. doi: 10.1016/j.jbi.2021.103977.
[17] Li T, Li N. On the tradeoff between privacy and utility in data publishing. In: Proceedings of the 15th ACM SIGKDD international conference on Knowledge discovery and data mining; 2009 Jun; Paris, France. New York: ACM; 2009. p. 517-526. doi: 10.1145/1557019.1557079.
[18] Prasser F, Kohlmayer F, Lautenschläger R, Kuhn KA. Arx-a comprehensive tool for anonymizing biomedical data. In: AMIA Annu Symp Proc. 2014;2014:984-993. American Medical Informatics Association; 2014.
[19] Cohen A, Nissim K. Towards formalizing the GDPR's notion of singling out. Proc Natl Acad Sci USA. 2020 Apr 15;117(15):8344-8352.